\documentclass[acmsmall]{acmart}
\usepackage{graphicx}

\usepackage{amsfonts,amsmath}

\usepackage{bm}
\usepackage[ruled,vlined]{algorithm2e}
\usepackage{multirow}

\AtBeginDocument{%
  \providecommand\BibTeX{{%
    \normalfont B\kern-0.5em{\scshape i\kern-0.25em b}\kern-0.8em\TeX}}}

\setcopyright{acmcopyright}
\copyrightyear{2020}
\acmYear{2020}
\acmDOI{10.1145/1122445.1122456}

\acmJournal{TALLIP}
\acmVolume{19}
\acmNumber{3}
\acmArticle{111}
\acmMonth{4}
\setcopyright{none}
\renewcommand\footnotetextcopyrightpermission[1]{}
\begin{document}

\title{Neural Unsupervised Semantic Role Labeling}

\author{Kashif Munir}
\email{kashifmunir92@sjtu.edu.cn}
\author{Hai Zhao}
\email{zhaohai@cs.sjtu.edu.cn}
\author{Zuchao Li}
\email{charlee@sjtu.edu.cn}
\affiliation{%
  \institution{Department of Computer Science and Engineering; Key Laboratory of Shanghai Education Commission for Intelligent Interaction and Cognitive Engineering; MoE Key Laboratory of Artificial Intelligence, AI Institute, Shanghai Jiao Tong University}
  \streetaddress{Dongchuan road 800, Minhang district}
  \city{Shanghai}
  \country{China}
  \postcode{201101}
}
\thanks{This work was supported in part by National Key Research, and Development Program of China under Grant 2017YFB0304100, in part by Key Projects of National Natural Science Foundation of China (U1836222, and 61733011), in part by Huawei-SJTU long term AI project, Cutting-edge Machine Reading Comprehension, and in part by
Language Model. (Corresponding author: Hai Zhao.)}

\renewcommand{\shortauthors}{Kashif et al.}

\begin{abstract}
The task of semantic role labeling (SRL) is dedicated to finding the predicate-argument structure. Previous works on SRL are mostly supervised and do not consider the difficulty in labeling each example which can be very expensive and time-consuming.
In this paper, we present the first neural unsupervised model for SRL. To decompose the task as two argument related subtasks, identification and clustering, we propose a pipeline that correspondingly consists of two neural modules. First, we train a neural model on two syntax-aware statistically developed rules.
The neural model gets the relevance signal for each token in a sentence, to feed into a BiLSTM, and then an adversarial layer for noise-adding and classifying simultaneously, thus enabling the model to learn the semantic structure of a sentence.
Then we propose another neural model for argument role clustering, which is done through clustering the learned argument embeddings biased towards their dependency relations.
Experiments on CoNLL-2009 English dataset demonstrate that our model outperforms previous state-of-the-art baseline in terms of non-neural models for argument identification and classification. 
\end{abstract}



\maketitle
\thispagestyle{empty}
\pagestyle{empty}
\section{Introduction}

Semantic Role Labeling (SRL) aims at elaborating the meaning of a sentence by modeling the structure between a predicate and its arguments in a sentence, which is usually described by answering the question “\textit{Who did what to whom?}”. The semantic role annotation describes the relationship between a specific predicate and its corresponding argument, which includes the performer of the action, the influence of the action, the time of the action, and so on ~\citep{gildea2002automatic}. The example below presents four sentences with SRL results in PropBank style.
\begin{itemize}
\item[a.] [\textit{Tim}]\textsubscript{\textit{A0}} \textbf{\textit{broke}} \textit{the} [\textit{glass}]\textsubscript{\textit{A1}} \textit{with a} [\textit{stone}]\textsubscript{\textit{A2}}.
\item[b.] \textit{The} [\textit{stone}]\textsubscript{\textit{A2}} \textbf{\textit{broke}} \textit{the} [\textit{glass}]\textsubscript{\textit{A1}}.
\item[c.] \textit{The} [\textit{glass}]\textsubscript{\textit{A1}} \textbf{\textit{broke}} [\textit{yesterday}]\textsubscript{\textit{AM-TMP}}.
\item[d.] \textit{One} [\textit{stone}]\textsubscript{\textit{A0}} \textbf{\textit{hit}} \textit{another} [\textit{stone}]\textsubscript{\textit{A1}}.  
\end{itemize}

The verb predicates are shown in boldface, and their arguments are shown in brackets. 
The subscripts are semantic roles of corresponding arguments. 
SRL is often constituted of two parts: \textit{argument identification} (e.g., \textit{Tim}, \textit{glass} and \textit{stone} in sentence a) and \textit{argument classification} (\textit{Tim} is A0, \textit{glass} is A1 and \textit{stone} is A2 for predicate \textit{broke} in sentence a).
SRL has wide range of applications in Natural Language Processing (NLP) including neural machine translation ~\citep{shi2016knowledge}, information extraction ~\citep{surdeanu2003using}, question answering ~\citep{berant2013semantic,yih2016value} etc.

Semantic role labeling can be categorized into two categories, span and dependency. Both types of SRL are useful for formal semantic representations but dependency based SRL is better for the convenience and effectiveness of semantic machine learning. \citet{johansson2008dependency} concluded that the best dependency based SRL system outperforms the best span based SRL system through gold syntactic structure transformation. The same conclusion was also verified by \citet{li2019dependency} through a solid empirical verification. Furthermore, since 2008, dependency based SRL has been more studied as compared to span based SRL. With this motivation, we intend to devise a strategy to perform dependency based SRL in an unsupervised manner.

Like other processes in NLP, the data-driven supervised model is an important choice in SRL.  These models promise good labeling results when there is already enough labeled data on the speciﬁc genre. However, when facing out-of-domain data, the $\mathrm{F_1}$-score will degrade over 10\%. 
\citet{marquez2008semantic} achieve a score of 81\% in supervised argument identification and 95\% in argument classification. \citet{pradhan2008towards} demonstrated that algorithm trained on Propbank ~\citep{palmer2005proposition} gives a decreased performance (approximately 10\%) on out of domain data.
Besides, performing SRL using a supervised model trained on one language but tested on another will further worsen the performance.
Furthermore, labeling data requires a lot of linguistic experts to provide resources, while unlabeled data is easy to collect in NLP. The dependency of algorithms on role-annotated data is causing hindrance for wide range applications of SRL.

In the recent past, a few papers focusing on the unsupervised method of SRL have been published as compared to the supervised ones. With the unavailability of annotated data, this task is very challenging to perform in an unsupervised manner. To our best knowledge, this can be verified by the fact that the last paper focusing on both unsupervised argument identification and classification was published in 2011 \cite{lang2011unsupervised,lang2011unsupervisedgraph}. However, there have been few papers focusing on semantic role classification solely published recently ~\citep{luan2016multiplicative,garg2012unsupervised,titov2014unsupervised}. This is the second paper inline after \citet{lang2011unsupervised}, focusing on both argument identification and classification in an unsupervised manner. Besides, even though neural models in terms of deep learning have been broadly applied in NLP, to our best knowledge, it is a bit surprising that this is the first neural modeling work handling the full unsupervised SRL task.

In this paper, we present an unsupervised model of SRL to identify and classify the arguments of verb predicates.
Following the same prerequisite as \citet{lang2011unsupervised}, our model only makes use of Part-of-speech (POS) tags and syntactic parse trees of sentences, which require comparatively less effort.

For argument identification, we train a neural model on two syntax-aware rules, which are developed via thorough statistical and unsupervised\footnote{the model does not make use of semantic annotations provided in CoNLL-2009 dataset but rather takes guidance from generic natured heuristic rules.} analysis of the English language. For each given sentence, the neural model feeds the relevance signal from each token in a sentence to BiLSTM and then to an adversarial layer for noise-adding.
The higher level relevance signal extracted by BiLSTM is aggregated by using a multi-layer perceptron and then a softmax layer is applied to predict the label of a token. However, it is quite possible that the model encodes a bias towards a particular category-related feature seen during the training (i.e., model overfitting over heuristic rules). To avoid this, we further introduce an adversarial layer over the relevance aggregation procedure to add noise in the form of a negative sample of a token label.
The addition of noise via the adversarial layer ensures that no category-specific information is memorized during the training, which in turn allows the model to generalize and surpass the upper-bound performance of simple heuristic rules.

For classification, we learn argument embeddings from context by explicitly incorporating dependency information and cluster the learned embeddings using an agglomerative clustering algorithm. The embeddings initialized with pretrained language model BERT~\citep{devlin-etal-2019-bert} are reduced to a lower dimension with the integration of an autoencoder approach.

\section{Related Work}
Semantic role labeling was first introduced by \citet{gildea2002automatic}, also known as shallow semantic parsing. With the introduction of deep learning, a series of neural SRL models have been proposed ~\citep{zhou-etal-2020-parsing,li-etal-2020-high,he2019syntax,li2019dependency,li2020memory,he2019syntaxx,che2008using,munir2021adaptive}. \citet{foland2015dependency} used convolutional and time-domain neural networks to form a semantic role labeler.
\citet{marcheggiani2017encoding} leveraged the graph convolutional network to incorporate syntax into a neural SRL model. \citet{yang2016learning} learn generalized feature vectors for arguments with a strong intuition that arguments occurring in the same syntactic positions bear the same syntactic roles.

Most of the work on SRL is supervised ~\citep{marquez2008semantic}, but there are also a majority of semi-supervised algorithms that focus on the task of SRL using fewer annotation resources. 
They learn the annotations from labeled data and project the same annotations to unseen data of the same language ~\citep{furstenau2009graph}.
\citet{pado2009cross} use the same concept and visualize the annotations and infer the results in another language. 
\citet{gordon2007generalizing} propose a semi-supervised model to observe the syntactically similar verbs within data and replicate these annotations to train the model.

Some authors have proposed unsupervised methods of argument identification and classification. 
\citet{abend2009unsupervised} utilize an unsupervised approach to identify arguments.
The idea is to find the minimal clause in a sentence containing the predicate in question and consider argument candidates are within the minimal clause.
In the second stage, they discard the weakly collocated argument-predicate pairs.

\citet{swier2004unsupervised} present the first method of unsupervised SRL.
This method does not use any annotated corpus. However, it uses Verb-Net ~\citep{kipper2000class}—a wide-ranging verb lexicon, which is very useful in identifying the possible argument structures for each verb.
Assembling Verb-Net along with rule-based chunker and a supervised version of syntactic parser ~\citep{swier2005exploiting}, the algorithm matches syntactic patterns in the corpus with Verb-Net.
Using it as a seed for the bootstrapping algorithm, the algorithm finds the verb arguments and classify them as per their semantic roles.

\citet{abend2010fully} classify the adjunct and core arguments by utilizing the POS tags and unsupervised parser for each sentence. 
\citet{lang2011unsupervisedgraph} view semantic role induction as a graph portioning method. The algorithm develops a weighted graph for each verb predicate, where the vertices denote the arguments. 
In the second stage, the graph partitioning creates clusters of vertices by using a variant of Chinese Whispers ~\citep{biemann2006chinese}.

\citet{grenager2006unsupervised} introduce a directed graph model by relating a verb predicate, its semantic roles, and their possible syntactic realizations.
The latent variables indicate the semantic roles, and these roles are classified by utilizing the states of these latent variables. 
\citet{levy2014dependency} and \citet{luan2016multiplicative} use the idea of infusing dependency relations in word embeddings to get word embeddings tightly focused on semantic relations.
Then they classify the arguments by performing agglomerative clustering on these word embeddings.

In recent years, neural models keep creating a new hype for supervised SRL ~\citep{roth2016neural,he2018syntax,li2018unified,foland2015dependency}, resulting in an increment of approximately 5\% performance score as compared to traditional methods. However, we have seen an improvement in bottleneck in very recent days. More importantly, no matter how good results are obtained on the test set of benchmark datasets, supervised SRL highly relies on the annotated dataset, and is stuck with the specific domain that its model is trained on. That is why we still need unsupervised SRL, even though it provides less satisfactory performance, but releases us from the dependency on annotated datasets and specific domains.

Following the line of \citet{abend2009unsupervised} and \citet{lang2011unsupervised}, we propose the first neural solution to the complete task of unsupervised SRL, argument identification and argument clustering.
We train a neural model on two statistically developed rules to identify arguments and then another neural model to learn embeddings of arguments biased towards their dependency relations. The learned embeddings are then clustered as per their semantic roles.

\section{Model}

The task of SRL is often performed in two steps, \textit{argument identification} and \textit{argument classification\footnote{Strictly speaking, a full SRL task actually includes four subtasks when taking the predicate identification and disambiguation into account. However, predicate-related evaluation weighs lighter among all SRL, and usually predicate related subtasks are much easier to obtain a good performance, which lets researchers exclude the predicate part from the unsupervised learning of SRL. Following \citet{lang2011unsupervised}, we keep this convention in this work by assuming predicates are known throughout this work.}}.
We perform both these tasks in an unsupervised manner.
Following the previous work \cite{lang2011unsupervised}, we only make use of POS tags and syntactic parse trees of sentences while identifying arguments.
The task of argument classification is considered as a clustering problem that clusters all the arguments of one category in a single group. 

Given a verb predicate and the dependency parse tree of a sentence, our model identifies the arguments in a sentence and assigns them their semantic roles.

\subsection{Argument Identification}
\label{neuralidentification}
\begin{figure}
\centering
  \includegraphics[width=0.95\linewidth,height=9.2cm]{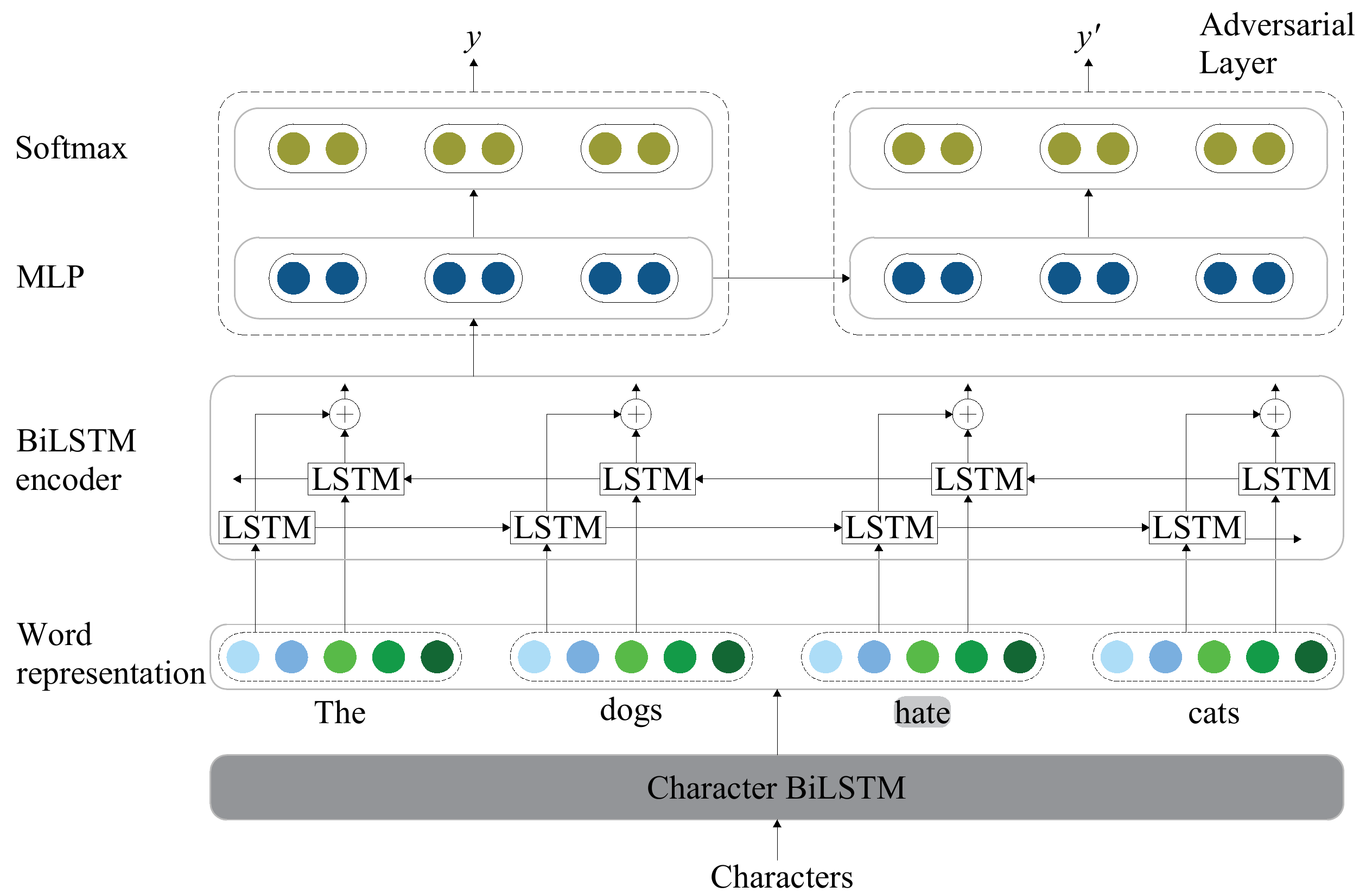}
  \caption{Argument identification model.}
  \label{fig:nn}
\end{figure} 
Figure~\ref{fig:nn} shows the illustrated network to identify arguments in a sentence. Following the previous convention ~\citep{marcheggiani2017encoding}, we consider predicate-specific word representation for a given sentence and a known predicate. Each word representation $x_i$ is formed by a concatenation of several features: 1) a randomly initialized embedding $x_i^r$. 2) a randomly initialized lemma embedding $x_i^l$. 3) character level embedding $x_i^{ch}$. 4) POS tag embedding $x_i^{POS}$. 5) a predicate related information feature $x_i^f$ which is typically a flag \{1,0\} indicating if a particular word is a predicate or not. The final word representation will be: $x_i = [x_i^r, x_i^l, x_i^{ch}, x_i^{POS}, x_i^f]$.
We apply bi-directional Long Short-Term Memory (LSTM) to the characters of a token, and the last hidden state of this BiLSTM generates the character level embeddings.
We further apply bi-directional LSTM to generate new representations of all the words in a sentence.
\begin{gather}
\label{bilstm}
\overleftarrow{w_t} = \overleftarrow{LSTM}(x_t, \overleftarrow{w_{t-1}}),\quad
\overrightarrow{w_t} = \overrightarrow{LSTM}(x_t, \overrightarrow{w_{t+1}}),\quad
 w_t = [\overleftarrow{w_t};\overrightarrow{w_t}]
\end{gather}
The hidden features in $w_t$ are further distilled through a fully connected layer.
\begin{equation}
h_{c,s} = g_a(W_1*w_t + b_1)
\end{equation}
where $W_1$ and $b_1$ are weight matrix and bias vector respectively. $g_a$ represents the ${\rm tanh}$ non-linearity function. The hidden feature $h_{c,s}$ is then fed to a binary classifier layer, which will produce a label $ \in \{0,1\}$ for each token in a sentence.
\begin{equation}
p(c|s) = softmax(W_2*h_{c,s} + b_2)
\end{equation}
where $W_2 $ and $b_2$ are the weight matrix and bias vector respectively.

\textbf{Adversarial Layer:} We expect the feature $h_{c,s}$ to be independent of the category. However, there is no guarantee that category related information is not infused in hidden feature $h_{c,s}$. We need to ensure that no category-specific information is memorized during the training otherwise the model will not be able to generalize and surpass the upper-bound performance of heuristic rules. We seek to achieve this by adding noise via an adversarial layer.
Specifically, we introduce a classifier over the hidden feature $h_{c,s}$ to find out the probability that $h_{c,s}$ belongs to each category seen during the training: $p'(.|h_{c,s}) = softmax(W_3*h_{c,s} + b_3)$.
The aim is to optimize the parameters $\phi = \{W_1, W_2, W_3, b_1, b_2, b_3\}$ so that each word in a sentence is classified to its true category. 
The overall loss of the model becomes:
\begin{equation}
    L( \phi) = \frac{1}{|N|} \sum_{(c,y) \in N} -p(y|h_{y,s}) + 
     \frac{\zeta}{|N|} \sum_{(c,y')\in N} -p'(y'|h_{y',s})
     \label{eq:temporal-difference}
\end{equation}

\noindent where $\zeta$ is a hyper-parameter, and $|N|$ is the number of training examples. $y$ are the silver target labels which we generate by using pruning and linguistic rules, thus making the model unsupervised. The generation of silver targets is explained in Section~\ref{silv}. $y'$ are the opposite of silver targets. The opposite targets are integrated into the loss function to create uncertainty in the model so our neural model can perform better than simple linguistic rules. 

\begin{algorithm}
\SetAlgoLined
\SetKwInOut{Input}{Input}\SetKwInOut{Output}{Output}
\Input{A predicate $p$, the dependency tree $T$, generic pairs $P$ and tuples $Tu$ as mentioned above}
\Output{The set of arguments $S$}
set $p$ as the current node $a$, $a=p$\\
\For{\upshape{each} descendant $n_i$ of $a$ in $T$}{
\If{$d(a,n_i) = 1$ \upshape{and} $n_i \notin S$}{
$CP = ({n_i}_{POS}, R_{n_i})$\\
\If{$CP$ \upshape{in} $P$}{
$S=S \cup n_i$}}}
set root $r$ as the current node $a$, $a=r$\\
\For{\upshape{each} descendant $n_i$ of $a$ in $T$}{
\If{$d(a,n_i)=d(a,p)$ \upshape{and} $n_i \notin S$}{
$CT = (R_p, R_{n_i}, {n_i}_{POS})$\\
\If{$CT$ \upshape{in} $Tu$}{
$S=S \cup n_i$
} 
}
}
\caption{\label{sivertargets} Silver targets generation}
\end{algorithm}

\subsection{Silver Targets Generation}
\label{silv}

We develop two heuristic rules using the syntax of English sentences to identify the arguments. Inspired by \citet{zhao2009semantic} and \textit{k-order} pruning algorithm of \citet{he2018syntax}, we present a model which searches through the entire parse tree of a sentence and matches the segments of a parse tree against the rules. 
The segments of a tree satisfying the rules contain the arguments.
The statistical research on the English language shows that: 

\noindent$\bullet$ when arguments are children of a predicate, there are 300 pairs of (\textit{argument POS tag, syntactic relation between predicate and argument}), which occur frequently. 

\noindent$\bullet$ when argument and predicate share the same syntactic head or the relative distances of predicate and argument from tree root is same, there are 20 tuples of (\textit{dependency relation of a predicate with its head, dependency relation of an argument with its head, POS tag of argument}) which occur frequently. 

The whole algorithm for silver targets generation is explained in Algorithm~\ref{sivertargets}. In the algorithm, $n_{POS}$ indicates the POS tag of node $n$, $R_n$ indicates the syntactic relation of node $n$ with its syntactic head and $d(n, n_d)$ indicates the syntactic distance between two nodes.
\begin{table}
\centering
\caption{\label{rules} Few examples of generic pairs and tuples used for silver targets generation. In algorithm, we use 300 pairs and 20 tuples.}
\begin{tabular}{|l|l| }
\hline
\textbf{Pairs $P$}   & \textbf{Tuples $Tu$} \\
\hline
(IN, ADV) & (VC, SBJ, NN)\\
\hline
(NNS, SBJ) & (ADV, SBJ, NNS)  \\
\hline
(NN, OBJ) & (VC, SBJ, PRP)\\
\hline
(IN, LGS)  &(VC, DEP, CC)  \\
\hline
(VBZ, OBJ)  & (VC, ADV, RB)\\
\hline
\end{tabular} 
\end{table}

\begin{figure}
\centering
  \includegraphics[width = 0.85\linewidth, height=3.3cm]{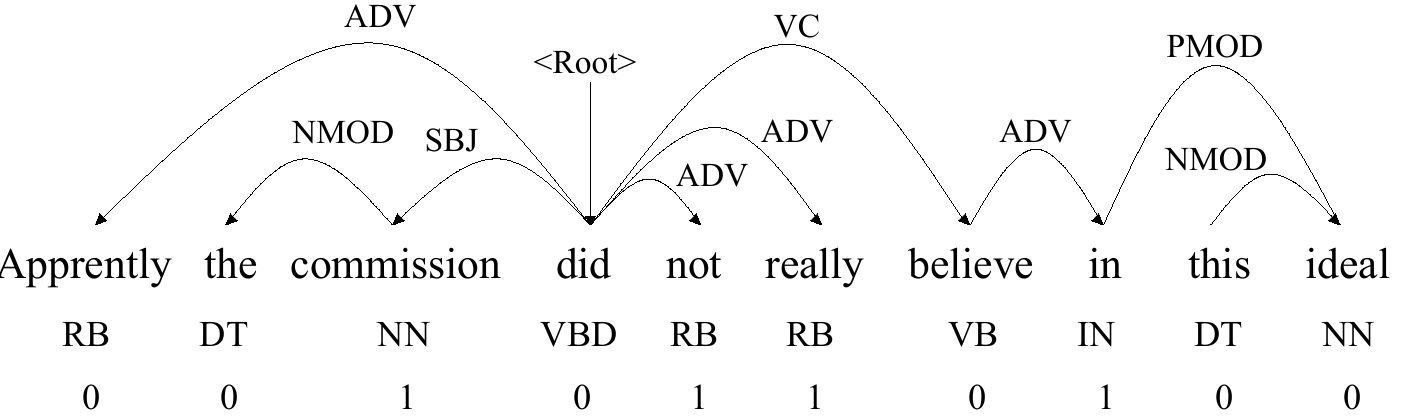}
  \caption{Example of a dependency parse tree. The dependency labels are shown above the arrows, while POS tags are shown below words. The last line shows which words are arguments for predicate \textit{believe}.}
  \label{fig:parsetree}
\end{figure}

We exemplify the rules by finding the arguments for predicate \textit{believe} in sentence \textit{‘Apparently the commission did not really believe in this ideal’}. 
The parse tree for this sentence is shown in Figure~\ref{fig:parsetree}. 
Stage 1  will prune the parse tree in a way that it contains only the children of a predicate. 
The parse tree for the sentence in consideration has only one child \textit{in} which will be identified as the argument because its pair (IN, ADV) is present amongst the frequent occurring 300 pairs as can be seen in Table~\ref{rules}. 
The next stage will appoint the argument candidates to be a set of \{\textit{really, apparently, commission, not}\} because their syntactic distances from root are same as the syntactic distance of predicate from root.
The tuple for \textit{really} becomes (VC, ADV, RB) which is a constituent of 20 generic tuples.
So, \textit{really} will be identified as an argument. 
Similarly, \textit{apparently}, \textit{commission} and \textit{not} are also arguments in this example.

\subsection{Argument Classification}
\label{argclass}
\begin{figure*}
  \includegraphics[width=1\textwidth,height=8.8cm]{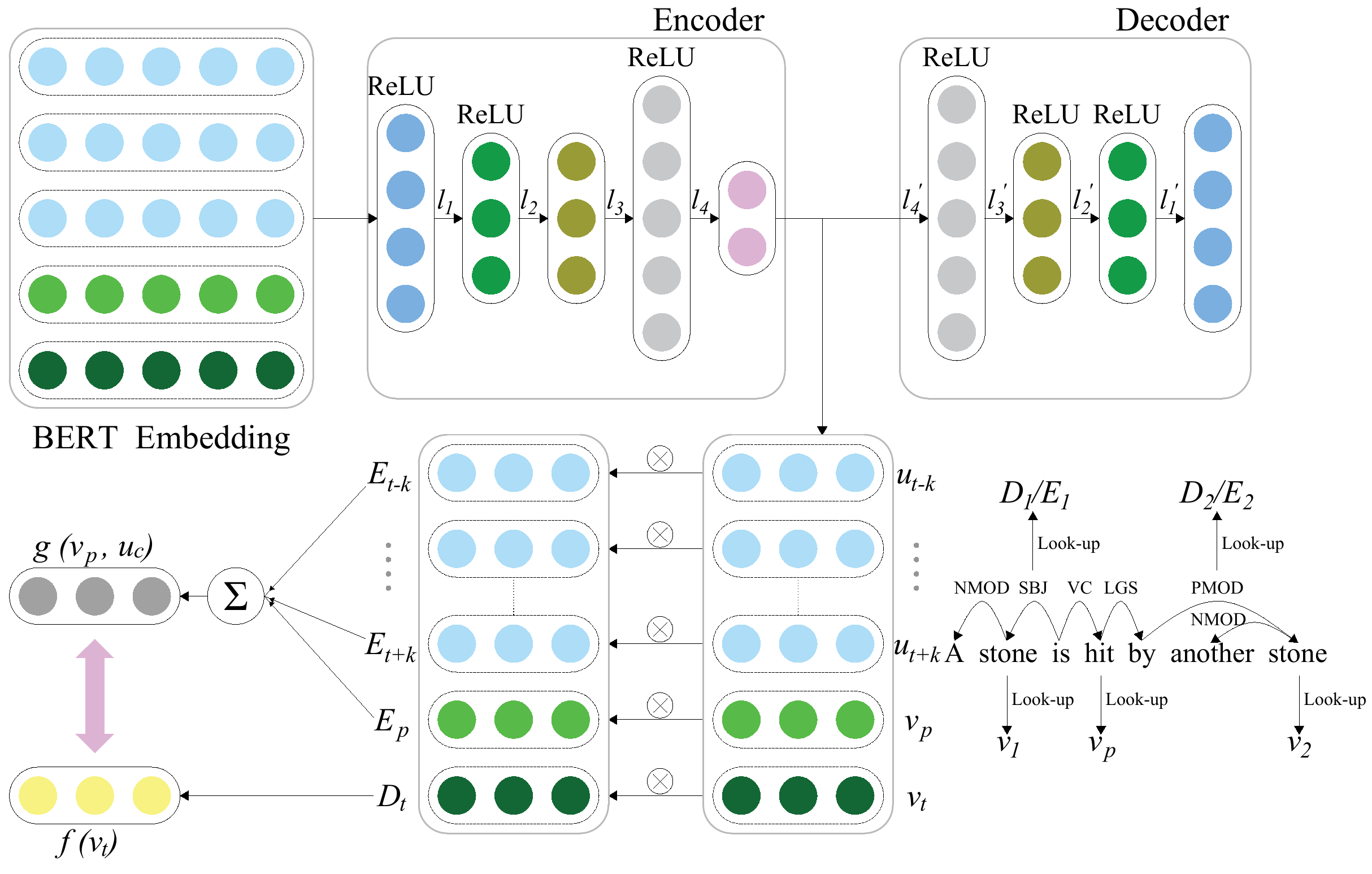}
  \caption{Overall model for learning argument embeddings. Autoencoder is used to reduce the embeddings dimension. $\otimes$ indicates $\textrm{tanh}$ operation. The biasing of dependency is uniformly applied to all argument embeddings.}
  \label{fig:autoencoder}
\end{figure*} 
Following the common practice handling unsupervised classification task, we devise argument classification as a clustering problem. 
The goal is to aggregate arguments of similar roles in the same clusters.
Following \citet{luan2016multiplicative}, we first generate embeddings for arguments that are tightly focused on relational semantics and feed these embeddings to the clustering algorithm.
We create a set of clusters for each verb predicate and then merge these clusters. 

\textbf{Representations:} Let $\mathit{v}_t$ be the embedding of the $t^{th}$ argument, $\mathit{v}_p$ be the embedding of the predicate and $u_c = \{u_{t-k},,,u_{t-1},u_{t+1},..,u_{t+k}\}$ are context vectors surrounding the $t^{th}$ argument with a window size $k$.
We initialize these embedding by using a pre-trained BERT model.

To facilitate the learning of clustering representations, we reduce the BERT embeddings dimensions by pre-training a stacked autoencoder (SAE). The structure of the autoencoder is shown in Figure ~\ref{fig:autoencoder}. 
The encoder encodes the embeddings to lower dimensions in the initial stage by using stacked fully connected layers, while the decoder maps low dimensional vectors into original dimension space using converse layers corresponding to the encoder.

The autoencoder is optimized using the least-square loss of input and output. The activation function in all layers except $l_4$ and $l'_1$ is $\textrm{ReLU}$. We do not use $\textrm{ReLU}$ for $l_4$ and $l'_1$ because we need $l_4$ to retain full information and layer $l'_1$ to be close to the input as much as possible ~\citep{vincent2010stacked}. We expect the output $l_4$ to retain most of the information from the preliminary embeddings of words. We first train the autoencoder to reduce the dimensions of $\mathit{v}_t$, $\mathit{v}_p$ and $u_c$. The next explained part in this section uses the reduced dimension version of embeddings.

Now the low dimension embeddings of arguments are further trained using the concept presented by \citet{luan2016multiplicative}. The prediction of the $t^{th}$ argument is:
\begin{equation}
\label{neural}
p(\mathit{v}_t | \mathit{v}_p , u_c) \propto {\rm exp}(\bm{f}(\mathit{v}_t)^\top \bm{g}(\mathit{v}_p , u_c))
\end{equation}
\begin{equation}
\bm{f}(\mathit{v}_t) = {\rm tanh}(\bm{D}_t \mathit{v}_t)
\end{equation}
\begin{equation}
\label{feedforward}
\bm{\mathit{g}}(\mathit{v}_p , u_c) = {\rm tanh} (\bm{E}_p \mathit{v}_p) + \sum_{u_i \in u_c} {\rm tanh}(\bm{E}_i u_i)
\end{equation}
where $\textrm{tanh}(.)$ is the element-wise $\textrm{tanh}$ function which helps in composing argument and its corresponding dependency relation with a non-linear multiplicative factor. This operation encourages the decoupling of dependency relations and arguments, which is useful in learning representations focused on lexical and relational semantics, respectively. $\bm{f}(.)$ and $\bm{\mathit{g}}(.)$ are the transformation functions for target argument embedding and context vectors respectively. The biasing effect for the dependency relation of a predicate with the $t^{th}$ argument is represented by $\bm{D}_t$. The first dependency relation on the path from argument's head to predicate is used as dependency label. $\bm{E}_p = \bm{I}$ because predicate has no relation with itself. $\bm{E}_i$ is the first dependency relation on the path from $i^{th}$ neighbor word to predicate. Equation~\ref{feedforward} represents a feed-forward neural network. To optimize the $softmax$ in the objective function, we use negative sampling as proposed by \citet{mikolov2013distributed}. Let $W_{neg}$ are negative samples for the the target argument embedding, the objective function becomes:
\begin{equation}
    J(\theta;\mathit{v}_t,u_c) = - \sum_{u_i \in u_c}  \log \sigma (u_i . \mathit{v}_t) - 
    \sum_{e_{neg} \in W_{neg}} \log \sigma (- e_{neg}.\mathit{v}_t)
\end{equation}

\textbf{Clustering:} The learned embeddings of arguments are then clustered by using agglomerative clustering. 
For each verb predicate, a number of seed clusters are created, which are merged hierarchically. The similarity between clusters is defined as cosine similarity of two clusters in the same way as proposed by \citet{luan2016multiplicative}.
\begin{equation}
    S(C,C') = cossim(x,y) -\alpha.pen(C,C')
\end{equation}
\begin{equation}
    pen(C,C') = \frac{|V(C,C')|+|V(C',C)|}{|C|+|C'|}
\end{equation}
where $|.|$ is cardinality of a set, $x$ and $y$ are centroids of clusters $C$ and $C'$ respectively and $\alpha$ is heuristically set to 0.1.

\section{Experiments}
\label{sec:length}

\begin{table*}[t]
\centering
\caption{\label{classification} SRL results with our proposed model, split-merge approach proposed in \citet{lang2011unsupervised} and syntactic function baseline which assigns arguments to clusters based on their syntactic functions.}
\begin{tabular}{lccccccccc}
\hline
\multirow{1}{*}{ }  & \multicolumn{3}{c}{Syntactic Function} & \multicolumn{3}{c}{Split-Merge} & \multicolumn{3}{c}{Ours}  \\ \cline{2-10} 
              & PU    & CO    & $\mathrm{F_1}$            & PU   & CO   & $\mathrm{F_1}$
              & PU    & CO    & $\mathrm{F_1}$ \\ \hline
auto/auto   & 72.9  & 73.9  & 73.4          & \textbf{81.9} & 71.2 & 76.2    & 81.3 & \textbf{74.8} & \textbf{77.9}     \\
gold/auto   & 77.7  & \textbf{80.1} & 78.9          & \textbf{84.0} & 74.4 & 78.9    & 83.7 & 78.8 & \textbf{81.2}\\
auto/gold   & 77.0  & 71.0  & 73.9          & 86.5 & 69.8 & 77.3    & \textbf{86.7} & \textbf{72.6} & \textbf{79.0} \\
gold/gold   & 81.6  & 77.5  & 79.5          & \textbf{88.7} & 73.0 & 80.1    & 85.2 & \textbf{79.4} & \textbf{82.2}\\

\hline \hline
\end{tabular}
\end{table*}


\begin{table*}[t]
\caption{\label{embedding} The result of using different word embedding techniques. }
\begin{tabular}{lcccccc}
\hline
\multicolumn{1}{c}{\multirow{2}{*}{System}} & \multicolumn{3}{c}{auto parses}          & \multicolumn{3}{c}{gold parses}                        \\ \cline{2-7} 
\multicolumn{1}{c}{}                        & PU             & CO             & $\mathrm{F_1}$    & PU             & CO             & $\mathrm{F_1}$             \\ \hline
Syntactic Function                            & 77.0          & 71.0 & 73.9 & 81.6           & 77.5           & 79.5           \\ 
Dependency-based embedding                     & \textbf{81.2} & \textbf{73.5}          & \textbf{77.2} & 84.2          & \textbf{78.8} & \textbf{81.4} \\ 
Linear-based embedding                        & 81.0          & 72.5          & 76.5 & \textbf{84.8} & 77.2          & 80.8          \\ \hline \hline
\end{tabular}

\end{table*}

\begin{table*}[t]
\caption{\label{ablation} Results for ablation study. NN ARGIDF represents the proposed neural model of argument identification, NN Clustering is the proposed model of semantic roles classification, Lang's ARGIDF denotes the argument identification rules proposed in \citet{lang2011unsupervised} and SYMDEP is argument classification algorithm as proposed in \citet{luan2016multiplicative}.}
\begin{tabular}{lccc}
\hline
System                      & PU    & CO    & $\mathrm{F_1}$    \\ \hline
NN ARGIDF + Split-Merge     & 81.2 & \textbf{79.0} & 80.1 \\ 
NN ARGIDF + NN Clustering   & 83.7 & 78.8 & \textbf{81.2} \\ 
Lang's ARGIDF + Split-Merge & 84.0  & 74.4  & 78.9  \\ 
NN ARGIDF + SYMDEP          & \textbf{84.2} & 77.9 & 80.9 \\ \hline \hline
\end{tabular}
\end{table*}

Both proposed identification\footnote{The code will be released at \url{https://github.com/kashifmunir92/Unsupervised_SRL}} and clustering models are tested on CoNLL-2009 English dataset. Following previous works ~\citep{abend2009unsupervised,marquez2008semantic}, we identify and classify the arguments for only verb predicates. 
CoNLL-2009 English dataset provides gold annotations in PropBank style. In addition to POS tags and gold dependency parses, it also provides the auto parses obtained by using MaltParser \cite{nivre2006maltparser}.
For argument identification, the dimension for  $x_i^r, x_i^l$ and $x_i^{ch}$ is 100, the dimension for $x_i^{POS}$ is 32 and for $x_i^f$ is 16.
The model explained in Section~\ref{neuralidentification} is trained four times for different values of $\zeta$, and then we ensemble these four models to identify arguments.
The number of training epochs is 24 and the batch size is 128.
Firstly, we increase $\zeta$ from 0.09 to 0.095 exponentially during the training to get the first model. 
Similarly, we increase $\zeta$ from 0.095 to 0.1, 0.1 to 0.105 and 0.105 to 0.11 to get three further models. 
All the increments in $\zeta$ are done exponentially during the training.
We ensemble these four models to predict arguments for verb predicates in sentences.

For classification, we initialize all embeddings by using pretrained BERT model. The dependency label for each argument is the first dependency relation on the path from argument to predicate. For example, in Figure ~\ref{fig:parsetree} the dependency label for \textit{commission} is $SBJ^{-1}$. To approximate the $softmax$ in the objective function, we use negative sampling as proposed by \citet{mikolov2013distributed}. 
The model is trained by using Adagrad ~\citep{duchi2011adaptive} with L2 regularization.

\subsection{Evaluation Metrics}
\label{metrics}

For identification, we report Precision, Recall and $\mathrm{F_1}$-score, which are defined in a standard way. 
For classification, we report Collocation, Purity and $\mathrm{F_1}$-scores.
Suppose $G_j$ represents the argument instances of $j^{th}$ gold class, $C_i$ represents the argument instances of the $i^{th}$ cluster, and $N$ represents the total argument instances. 
Purity, Collocation and $\mathrm{F_1}$-score are defined as follows: 
\begin{gather}
PU = \frac{1}{N}\sum_{i} \max{j} |G_j \cap C_i|,\nonumber\\
CO = \frac{1}{N}\sum_{j} \max{i} |G_j \cap C_i|,\nonumber\\
\mathrm{F_1} = \frac{2 \times PU \times CO}{PU + CO}
\end{gather} 

\subsection{Baselines}
\label{sect:pdf}

We compare our model against the unsupervised SRL algorithm proposed in \citet{lang2011unsupervised}, which identifies arguments and then clusters them as per their semantic roles. The system uses a set of linguistically-motivated rules to identify arguments and then formulates the classification as a clustering problem. The clustering algorithm executes a series of split and merge phases, eventually transducing an initial clustering of high purity and less collocation to a final clustering of better quality. 
The second baseline for comparison is based on a syntactic function, which uses a parser to develop a syntactic function (e.g., subject, object) for each argument and assigns semantic roles to argument instances based on this syntactic function. This baseline has been used previously by \citet{lang2011unsupervised} and \citet{grenager2006unsupervised} and shown quite challenging to outperform.

\section{Results and Analysis}
\label{sec:length}
\begin{figure}
  \includegraphics[width=0.7\linewidth, height =5cm]{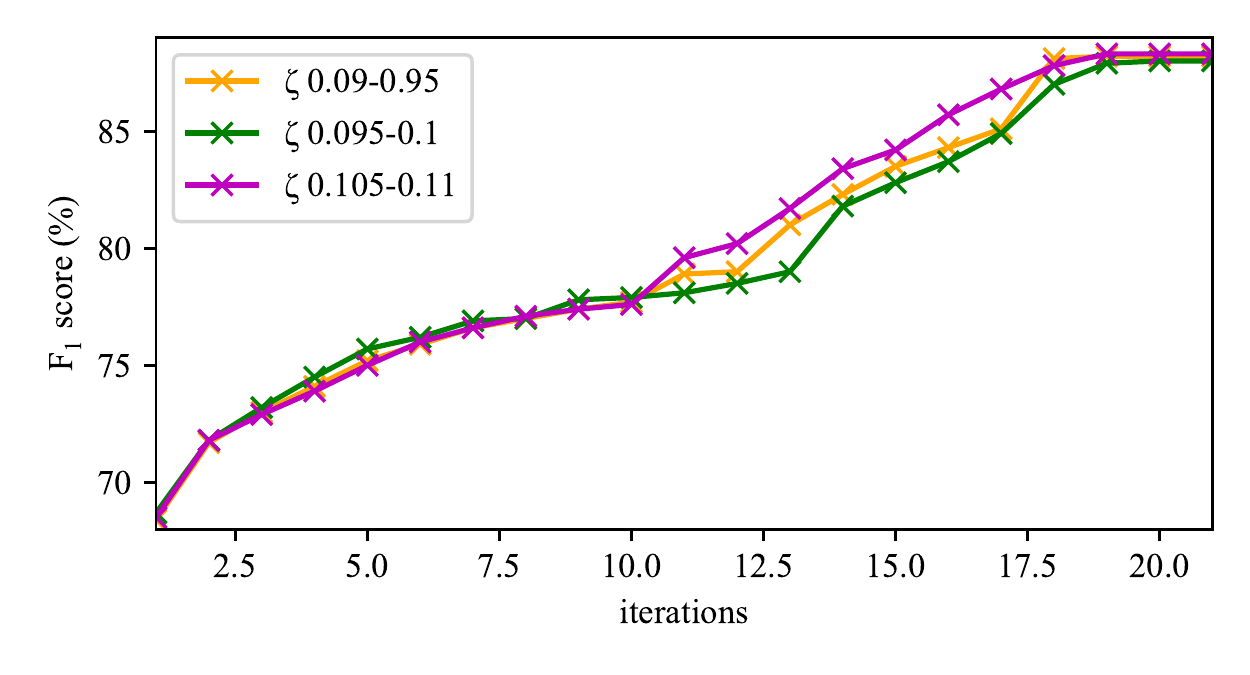}
  \caption{$\mathrm{F_1}$-score for argument identification against the number of iterations for different values of $\zeta$.}
  \label{fig:trainplot}
  
\end{figure}

The overall results of our unsupervised SRL system are shown in Table~\ref{classification}.
Originally, \citet{lang2011unsupervised} reported their results for CoNLL-2008\footnote{CoNLL-2008 task is only for English, while CoNLL-2009 is a multilingual task. The main difference is that predicates are pre-identified for the latter.} dataset, but here we report their model’s scores for CoNLL-2009 dataset.
Following \citet{lang2011unsupervised}, we compare our model against split-merge algorithm~\citep{lang2011unsupervised} and syntactic function baseline on four datasets. 

\noindent$\bullet$ combination of automatic parses and automatically identified arguments (auto/auto).

\noindent$\bullet$ combination of gold parses and automatically identified arguments (gold/auto).

\noindent$\bullet$ combination of automatic parses and gold arguments (auto/gold).

\noindent$\bullet$ combination of gold parses and gold arguments (gold/gold).

On all datasets, our model achieves the highest $\mathrm{F_1}$-score as compared to the state of the art system (split-merge) which has been a top-scoring unsupervised SRL system since 2011, which reflects that performing SRL in an unsupervised manner is quite challenging. On auto/auto dataset, our model results in 3.6\% higher collocation than split-merge and improves  $\mathrm{F_1}$-score by a margin of 1.7\%. Similarly, our models improve over split-merge algorithm when using gold parses. On gold/auto dataset, our model yields 2.3\% higher $\mathrm{F_1}$-score as compared to the split-merge algorithm. Not unexpectedly, our model also surpasses the baselines when gold arguments are used instead of automatically identified arguments.

To assess the performance of the argument identification part, we first generate silver targets using the Algorithm~\ref{sivertargets} and then train the model as mentioned in Section~\ref{neuralidentification}. For gold parses, the argument identification model identifies arguments with a precision of 94.0\% and a recall of 83.2\% as compared to the precision of 88.1\% and a recall of 87.9\% of \citet{lang2011unsupervised}. For auto parses, the argument identification model identifies arguments with a precision of 92.3\% and a recall of 82.9\% as compared to the precision of 87.2\% and a recall of 86.8\% of \citet{lang2011unsupervised}.

Compared to \citet{lang2011unsupervised} method, the most significant improvement of our method is that it uses word vectors focused on semantic relations for clustering.
Lang and Lapata’s method only measures the word distribution similarity of two clusters in essence. This method will perform worse on a smaller dataset, because words with similar meanings may not necessarily be similar semantically \cite{ashraf2018measures}.

To illustrate the necessity of training word vectors based on syntax, we train traditional word vector models based on linear window context and dependency syntax respectively under the same training conditions (English Wikipedia backup file in September 2013, about 5GB). 
The training parameters are the same -- the word vector dimension is five, the number of negative samples is five, and the number of iterations is two.
The window size of word embedding based on linear window context is set to two, and the CoreNLP Stanford syntactic parser is used for data preprocessing.
The learned embeddings are then clustered as per their semantic roles by using the split-merge algorithm (using gold arguments in CoNLL-2009).

As shown in Table~\ref{embedding}, word vectors based on dependency syntax perform better than the traditional word vectors based on linear window context in all aspects. $\mathrm{F_1}$-score for the original split-merge algorithm on gold/gold dataset is 80.1\%; however if we use dependency based embeddings to perform clustering by split-merge algorithm, $\mathrm{F_1}$-score on gold/gold dataset improves to 81.4\% which shows the strength of our approach.
This 1.3\% improvement is because dependency-based word vectors can give higher similarity scores to words with similar functions, while traditional word vectors can give higher similarity scores to words with similar topics. 
For example, among the word vectors we train, we search for the words most similar to \textit{students}.
The word vectors based on linear context will give \textit{straight-A} and \textit{public school}, while the word vectors based on dependency syntax will give \textit{undergraduate} and \textit{graduate}. 

Table~\ref{verbmalt} and~\ref{roles} show the comparison of our model against \citet{lang2011unsupervised} across various verb predicates and roles respectively on auto/auto dataset. The bold type in the table represents the better experimental results under the comprehensive index. Our model exceeds the baseline in terms of $\mathrm{F_1}$-score for most of the predicates and roles.

Finally, Figure ~\ref{fig:trainplot} shows the $\mathrm{F_1}$-score for argument identification by the proposed neural model against the number of iterations. 
\begin{table*}[t]
\centering
\caption{\label{verbmalt} Results of clustering for individual verbs in CoNLL-2009.}
\begin{tabular}{lccccccc}
\hline
\multirow{2}{*}{Verbs} & \multirow{2}{*}{Freq} & \multicolumn{3}{c}{Split-Merge} & \multicolumn{3}{c}{Ours}   \\ \cline{3-8} 
                       &                       & PU    & CO    & $\mathrm{F_1}$            & PU   & CO   & $\mathrm{F_1}$            \\ \hline
\textit{say}           & 15238                 & 93.6  & 81.7  & 87.2 & 93.8 & 84.8 & \textbf{89.1}          \\
\textit{make}          & 4250                  & 73.3  & 72.9  & 73.1          & 75.5 & 73.2 & \textbf{74.3} \\
\textit{go}            & 2109                  & 52.7  & 51.9  & 52.3          & 53.2 & 62.1 & \textbf{57.3} \\ 
\textit{increase}      & 1392                 & 68.8  & 71.4  & 70.1          & 75.7 & 74.7 & \textbf{75.2} \\
\textit{know}          & 983                  & 63.7  & 65.9  & 64.8          & 68.1 & 67.7 & \textbf{67.9} \\
\textit{tell}          & 911                  & 77.5  & 70.8  & 74.0          & 75.3 & 77.1 & \textbf{76.2} \\
\textit{consider}      & 753                  & 79.2  & 61.6  & 69.3          & 71.1 & 72.5 & \textbf{71.8} \\
\textit{acquire}       & 704                  & 80.1  & 76.6  & 78.3          & 80.3 & 79.5 & \textbf{79.9} \\
\textit{meet}          & 574                  & 88.0  & 69.7  & 77.8          & 80.7 & 79.7 & \textbf{80.2} \\
\textit{send}          & 506                  & 83.6  & 65.8  & 73.6          & 70.1 & 67.7 & \textbf{68.9} \\
\textit{open}          & 482                  & 77.6  & 62.2  & 69.1          & 71.9 & 72.7 & \textbf{72.3} \\
\textit{break}         & 246                  & 68.7  & 53.3  & 60.0          & 65.2 & 64.8 & \textbf{65.0} \\
\hline \hline
\end{tabular}

\end{table*}


\begin{table*}[t]
\centering
\caption{\label{roles} Results for clustering of individual semantic roles with our method and split-merge baseline.}
\begin{tabular}{lcccccc}
\hline
\multirow{1}{*}{Role}  & \multicolumn{3}{c}{Split-Merge} & \multicolumn{3}{c}{Ours}   \\ \cline{2-7} 
              & PU    & CO    & $\mathrm{F_1}$            & PU   & CO   & $\mathrm{F_1}$            \\ \hline
A0                            & 79.0  & 88.7  & 83.6          & 84.0 & 86.4 & \textbf{85.2}\\
A1                            & 87.1  & 73.0  & 79.4          & 79.9 & 80.3 & \textbf{80.1} \\
A2                            & 82.8  & 66.2  & \textbf{73.6}          & 75.3 & 70.6 & 72.9 \\ 
A3                            & 79.6  & 76.3  & 77.9          & 77.8 & 78.6 & \textbf{78.2} \\
ADV                           & 78.8  & 37.3  & 50.6          & 68.9 & 48.9 & \textbf{57.2} \\
CAU                           & 84.8  & 67.2  & \textbf{75.0}          & 78.1 & 70.9 & 74.3 \\
DIR                           & 71.0  & 50.7  & 59.1          & 71.1 & 52.3 & \textbf{60.3} \\
EXT                           & 90.4  & 87.2  & 88.8          & 90.0 & 90.2 & \textbf{90.1} \\
LOC                           & 82.6  & 56.7  & \textbf{67.3}          & 66.9 & 65.9 & 66.4 \\
MNR                           & 81.5  & 44.1  & 57.2          & 68.5 & 51.0 & \textbf{58.5} \\
TMP                           & 80.1  & 38.7  & 52.2          & 65.8 & 44.4 & \textbf{53.0} \\
MOD                           & 90.4  & 89.6  & 90.0          & 90.6 & 92.2 & \textbf{91.4} \\
NEG                           & 49.6  & 98.8  & 66.1          & 63.0 & 71.5 & \textbf{67.0} \\
DIS                           & 62.2  & 75.4  & 68.2          & 68.6 & 69.4 & \textbf{69.0} \\
\hline \hline
\end{tabular}
\end{table*}


\begin{table*}[t]
\caption{\label{adversarial} Comparison of argument identification model with and without adversarial layer with heuristic rules.}
\begin{tabular}{lcccccc}
\hline
\multicolumn{1}{c}{\multirow{2}{*}{System}} & \multicolumn{3}{c}{auto parses}          & \multicolumn{3}{c}{gold parses}                        \\ \cline{2-7} 
\multicolumn{1}{c}{}                        & P             & R             & $\mathrm{F_1}$    & P             & R             & $\mathrm{F_1}$             \\ \hline
Heuristic Rules           & 90.1 & 79.6 & 84.5 & 91.0 & 80.1 & 85.2   \\ 
w/o Adversarial Layer     & 90.3 & 78.9 & 84.2 & 90.5 & 79.6 & 84.7 \\ 
with Adversarial Layer    & \textbf{92.3} & \textbf{82.9} & \textbf{87.3} & \textbf{94.0} & \textbf{83.2} & \textbf{88.3}          \\ \hline \hline
\end{tabular}
\end{table*}

\subsection{Ablation Study}

To analyze the proposed unsupervised SRL model, we perform a series of ablation studies. For all the ablation experiments, we use gold parses and automatically identified arguments unless specified.

We create an ablated study to explore which part of our method performs better. The results of the ablation study are shown in Table ~\ref{ablation}.  While using split merge ~\citep{lang2011unsupervised} for clustering, we compare our argument identification model with \citet{lang2011unsupervised}. The results validate that our argument identification model surpasses \citet{lang2011unsupervised} by a margin of 1.2\%. Similarly, we keep the argument identification component constant and change the clustering component of the system. By using our argument identification, we cluster arguments by using split-merge, SYMDEP ~\citep{luan2016multiplicative} and our proposed clustering algorithm. The results show that our classification component outperforms the split-merge algorithm by a margin of 1.1\% and SYMDEP by a margin of 0.3\%.

To analyze the effect of using BERT in the argument classification task, we replace the pre-trained BERT model (along with autoencoder, explained in Section~\ref{argclass}) with BiLSTM model and train the unsupervised neural argument classification model in an end-to-end fashion to bias embeddings according to their dependency roles. This setting degrades $\mathrm{F_1}$-score to 80.1\% on gold/auto dataset which is still better than split-merge algorithm (78.9\% $\mathrm{F_1}$). This result confirms that features provided by pre-trained BERT embeddings help in the overall performance gain because BERT can offer more accurate context information as compared to BiLSTM. 

To investigate the advantage of adding an adversarial layer, we remove this layer and report the score of the argument identification model in Table~\ref{adversarial}. We also report scores for simple heuristic rules. The results show that the model without the adversarial layer performs worse than heuristic rules. However, the incorporation of the adversarial layer allows the model to improve argument identification performance by 3.1\% when using gold parses and 2.8\% when using auto parses. This validates that the addition of noise via the adversarial layer avoids the model to depend solely on heuristic rules and generalizes the ability of the model to learn the semantic structure of a sentence.

\section{Conclusion}
This paper presents a new approach towards unsupervised SRL. By incorporating syntax information, the proposed neural model can identify arguments for verb predicates without relying on extensive amount of training data. 
Then we propose another neural model to learn embeddings of arguments biased towards their dependency relations. These embeddings are then clustered as per their semantic roles. The proposed SRL model holds promise for reducing the data acquisition bottleneck for supervised systems.

\bibliographystyle{ACM-Reference-Format}
\bibliography{sample-acmsmall}

\end{document}